\documentclass[journal]{IEEEtran}
\usepackage[T1]{fontenc}
\usepackage[utf8]{inputenc}
\usepackage{amsmath,amssymb,amsfonts}
\usepackage{graphicx}
\usepackage{booktabs}
\usepackage{array}
\usepackage{cite}
\usepackage{url}
\usepackage[colorlinks=true,allcolors=blue]{hyperref}

\begin{document}

\title{Long-Horizon Causal Inertial Odometry in Multistory Buildings}

\author{Tianzi Zhang, Yan Wang, Hao Qiao, Jian Kuang, and Xiaoji Niu%
\thanks{Yan Wang is the corresponding author (e-mail: wystephen@whu.edu.cn).}%
\thanks{T. Zhang and H. Qiao are with the School of Geodesy and Geomatics, Wuhan University, Wuhan 430079, China. Y. Wang, J. Kuang, and X. Niu are with the GNSS Research Center, Wuhan University, Wuhan 430079, China.}}

\hypersetup{pdftitle={Long-Horizon Causal Inertial Odometry in Multistory Buildings},
pdfauthor={Tianzi Zhang, Yan Wang, Hao Qiao, Jian Kuang, Xiaoji Niu}}
\maketitle

\begin{abstract}
In indoor 3D localization inside multistory buildings, an elevator or an escalator accelerates for a few seconds and then cruises at constant velocity for tens of seconds. Gait excitation disappears, the specific force degenerates into a constant gravity reaction, and instantaneous velocity becomes unobservable within a 1--2 s window. The output of short-window learned inertial odometry consequently degrades toward the conditional mean of the training set, and vertical displacement is systematically underestimated. We propose Local-Token Causal Attention (LTCA), a long-horizon architecture that compresses each 1 s inertial window into a 1 Hz motion token and sustains, over a bounded 60 s causal cache, the motion state established at the onset event. We further introduce a pluggable barometer branch that, when online, supplies a direct elevation constraint to the vertical channel through local differential height observations, and we train it with paired online/fallback conditional training so that the barometer also regularizes the representation shared by the two modes during training. On MSIO, our benchmark of nine buildings, the RoNIN and TLIO instances reach 3D trajectory errors (ATE-3D) of 2.83/4.18 m and 2.44/5.12 m on the Seen and Unseen splits, respectively, with the barometer online. With the barometer disabled at inference, the same weights fall back to pure inertial inference and reach 3.24/4.28 m and 2.64/5.23 m, still below every native short-window baseline. We further validate the framework on the public RoNIN and TLIO benchmarks.
\end{abstract}


\section{Introduction}
\label{sec:1}

\IEEEPARstart{I}{n} GNSS-denied indoor and underground environments, localizing a pedestrian inside a multistory building requires estimating the horizontal trajectory and the inter-floor elevation jointly. Visual-inertial odometry \cite{b1} is constrained by illumination, texture, and power budget. Without external observations, a micro-electromechanical inertial measurement unit (IMU) accumulates integration error rapidly \cite{b2}, and the vertical channel is the worst affected. Conventional pedestrian dead reckoning relies on zero-velocity updates at the foot \cite{b3} or on empirical step-length models, and both usually assume a rigidly attached device. Once the device is carried in the hand, in a pocket, or in a backpack, these assumptions no longer hold. Obtaining a continuous cross-floor 3D trajectory from a consumer-grade IMU and barometer alone therefore remains an open problem in indoor localization.

Learned inertial odometry, represented by RoNIN \cite{b4} and TLIO \cite{b5}, regresses instantaneous velocity or displacement directly from high-rate inertial measurements, removing hand-crafted gait rules and mounting constraints. Its physical basis is the periodic foot--ground impact of walking: the impact waveform and the translational velocity share a stable statistical relation, so that a 1--2 s window suffices to map one onto the other. Elevators and escalators break this premise. After a brief acceleration, the transport platform enters a constant-velocity cruise that lasts tens of seconds; the net mechanical acceleration vanishes, the specific force degenerates into a constant gravity reaction, and the local signal becomes indistinguishable from standing still. The current window carries no information about instantaneous velocity, so a window-based estimator has nothing left to condition on. A stateless network with a 1--2 s receptive field possesses no cross-segment history, so its output collapses toward the conditional mean and vertical displacement is systematically underestimated. Recovering velocity therefore requires establishing causal state continuity across the excitation-free gap.

To address this problem, we propose a long-horizon architecture termed Local-Token Causal Attention (LTCA). At its bottom level, a 1 s bounded receptive field compresses the high-rate inertial stream into 1 Hz motion tokens while preserving the residual cancellation property of feed-forward convolution. At its top level, a learnable relative temporal bias is applied over a bounded 60 s causal cache, so that the motion state established at the onset event can be called upon throughout the cruise, and a strict causal mask supports 1 Hz streaming inference. What matters here is not a larger parameter count but keeping the onset event inside the visible history. LTCA introduces no independent absolute observation for the vertical channel, so vertical error still accumulates once an external elevation reference is absent. We therefore attach a pluggable barometer on top of it. At the structural level, identity pass-through and hard gating let one set of weights switch between online and fallback modes through a mask; at the training level, paired conditional training makes the barometer act as a privileged observation that shapes the shared representation, so that the fallback path retains most of the gain.

Compared with prior work, our main line is to restore long-horizon causal state with LTCA and to attach the barometer as a pluggable auxiliary observation within the same framework. Short-window learned inertial odometry \cite{b4,b5,b6,b7,b8,b9} relies on bounded receptive fields to cancel residual errors statistically, but cannot cover the onset event. IDOL \cite{b10} extends the training window to 20 s with curriculum learning, yet its bidirectional structure requires the whole window at inference and does not admit strictly causal streaming recursion. Learned multi-sensor fusion \cite{b11,b12} mostly assumes that the auxiliary sensors remain available throughout and provides no tolerance for packet loss or low-power sleep. Learning using privileged information (LUPI) \cite{b13,b14}, together with privileged geometric constraints in robot proprioception \cite{b15}, shows that training-time observations can shape the representation, but this idea has not yet been combined with the training protocol of pedestrian 3D odometry.

The contributions of this paper are threefold.

\begin{enumerate}
\item For continuous 3D localization involving elevators and escalators in multistory buildings, we attach the LTCA long-horizon structure and a pluggable barometric fusion module to two existing local backbones, RoNIN and TLIO, through their existing interfaces, extending learned inertial odometry from level walking to multistory scenes.
\item We explain why short-window learned inertial odometry fails during the constant-velocity cruise of passive mechanical transport: once gait excitation vanishes, the inertial window carries no information about velocity and the estimate falls back to its prior. We further identify the role of causal temporal attention in maintaining motion-state continuity across the excitation-free gap.
\item We build MSIO, a 3D multistory indoor localization benchmark rich in elevator data, carry out a unified cold-start comparative evaluation on MSIO and on the public RoNIN and TLIO benchmarks, and isolate the role of each component through per-motion-mode and ablation studies.
\end{enumerate}

\section{Related Work}
\label{sec:2}

\subsection{Learned Inertial Odometry and the Short-Window Paradigm}
\label{sec:2.1}

Conventional pedestrian dead reckoning relies on gait detection and hand-designed kinematic models, such as zero-velocity updates based on foot--ground contact characteristics \cite{b3} and cadence-based empirical step-length regression \cite{b2}. Data-driven methods instead learn the mapping from inertial windows to motion states. RIDI \cite{b16} recognizes the device carrying mode and regresses a correction to the body-frame acceleration. IONet \cite{b17} uses a bidirectional recurrent network to regress displacement and heading increments in polar coordinates from a sequence of slices. RoNIN \cite{b4} compares convolutional and recurrent structures on a large dataset in a gravity-aligned frame and shows that feed-forward convolution cancels residuals statistically in open-loop integration, which yields lower drift. TLIO \cite{b5} extends the regression to 3D relative displacement with heteroscedastic measurement covariance, and tightly couples the network with an extended Kalman filter that online-corrects the system state and the sensor biases.

Recent work introduces geometric priors and local inductive biases. CTIN \cite{b6} captures feature correlations with contextual self-attention inside a local time window. EqNIO \cite{b7} builds an equivariant convolutional network from invariance to rotations about the gravity axis. iMoT \cite{b8} adopts a two-branch structure to extract angular-rate and linear-acceleration features separately. X-IONet \cite{b9} explores adaptive alignment of motion features across platforms. CTIN, EqNIO, and iMoT are built on a local, stateless 1--2 s sliding window, and their validity rests on the presence of continuous biomechanical excitation within that window; X-IONet couples intra-sequence attention with an extended Kalman filter, but its displacement is still regressed from a bounded input sequence. In passive mechanical transport such as elevators and escalators, foot impact disappears, the motion acceleration drops to zero, and the specific force degenerates into a constant gravity reaction; the window carries no information about the velocity state, and the short-window output collapses toward zero velocity.

\subsection{Temporal Windows and Context Modeling in Neural Inertial Navigation}
\label{sec:2.2}

Mainstream work such as IONet \cite{b17}, RoNIN \cite{b4}, and TLIO \cite{b5} adopts local 1--2 s windows. RoNIN compares convolutional and recurrent networks under different window lengths. For a one-dimensional residual network, a 1--2 s window already covers a complete active gait cycle. The receptive field of feed-forward convolution is bounded, the estimation residuals at different time steps are approximately statistically independent, and positive and negative errors cancel in open-loop displacement integration, so that accumulated drift is suppressed. Restricting the temporal receptive field is the common principle by which subsequent local models (CTIN \cite{b6}, EqNIO \cite{b7}) keep open-loop drift under control. A local short window, however, cannot cover long-range history, and the model can hardly perceive macroscopic changes of motion state beyond the 2 s scale.

Several studies attempt to extend the temporal context. IDOL \cite{b10} uses a two-layer bidirectional long short-term memory network and a curriculum that expands the input sequence from 1 s to 20 s. To suppress the drift caused by long-range recurrent propagation, IDOL resets the hidden state at the window boundary during inference, so the hidden state does not propagate continuously across adjacent 20 s sequences. Its bidirectional structure and 20 s window require the whole window at inference and do not provide causal state recursion; following the official deployment configuration, we evaluate it with a 1 s window.

The design of temporal context involves a trade-off. A local short window keeps residuals independent, but its receptive field is small and it can only see the current state. Feeding tens of seconds of raw high-rate measurements directly incurs algorithmic latency and discontinuities of state at window boundaries. An unconstrained recurrent hidden state accumulates error autocorrelation under open-loop integration. A streaming temporal architecture that combines the independence of local features with macroscopic causal continuity therefore remains a direction worth exploring.

\subsection{Learned Multi-Sensor Fusion and Learning Using Privileged Information}
\label{sec:2.3}

Atmospheric pressure decays exponentially with altitude, so a consumer barometer provides a direct absolute elevation observation and is widely used to suppress the double-integration divergence of the inertial vertical channel. TinyOdom \cite{b12} keeps the barometer outside the network on a low-power embedded platform, running a separate second-order state estimator without learned parameters. MARIO \cite{b11}, targeting head-mounted devices, extracts features from dual inertial units, a magnetometer, and a barometer with causal convolution and concatenates them directly at a bottleneck layer. Most existing learned fusion schemes assume that the auxiliary sensors stay online throughout, and tolerate neither dynamic packet loss on mobile devices, bus congestion, nor low-power sleep. Decoupled external filtering, in turn, does not exploit the joint cross-modal representation.

Auxiliary multimodal observations can also shape the representation during training, and their effect goes beyond direct feature input. Learning using privileged information (LUPI) \cite{b13,b14} states that privileged observations, available only to the teacher side during training, constrain the capacity of the hidden manifold and steer the network toward causal, invariant features that generalize better. In robot autonomous navigation and proprioceptive state estimation \cite{b15}, simulated or paired high-precision observations serve as privileged geometric information and constrain onboard policies that otherwise rely on inertial history alone. For pedestrian indoor 3D navigation, a viable route is to provide an elevation constraint while the sensor is online and to fall back losslessly to pure inertial inference when it is not. Paired training uses the elevation difference as a privileged geometric cue to regularize the backbone representation, which improves the stability of the model.

\section{Method}
\label{sec:3}

\subsection{Problem Definition}
\label{sec:3.1}

Three-dimensional pedestrian inertial odometry operates under a causal-inference constraint: it may exploit only the current and past measurements, and it estimates the pedestrian's relative displacement increment. In the complete architecture, LTCA and the pluggable barometer share one coordinate convention and one integration protocol. Let $f_s$ denote the sampling rate and $\Delta t = 1/f_s$ the sampling period; let $T_o$ be the time interval between two consecutive displacement outputs, so that a single output step contains $N = f_s T_o$ samples, with $N = 100$ at the 100 Hz rate used in the experiments. Throughout, $C$ denotes the width of the backbone feature maps, $d$ the token dimension ($d = 128$), $H$ the number of attention heads with $d_k = d/H$, and $L = 60$ the number of tokens held in the causal cache.

Raw inertial measurements are produced in the body frame ($b$-frame) rigidly attached to the sensor. An onboard attitude estimator provides the attitude quaternion $\mathbf{q}_{t,k}$, which transforms the raw triaxial angular rate $\boldsymbol{\omega}_{t,k}^{b}$ and the specific force reading $\mathbf{f}_{t,k}^{b}$ into a local gravity-aligned navigation frame ($n$-frame) \cite{b4,b5}:

\begin{equation}
\boldsymbol{\omega}_{t,k}^{n} = \mathbf{q}_{t,k} \otimes \boldsymbol{\omega}_{t,k}^{b} \otimes \mathbf{q}_{t,k}^{-1},
\qquad
\mathbf{f}_{t,k}^{n} = \mathbf{q}_{t,k} \otimes \mathbf{f}_{t,k}^{b} \otimes \mathbf{q}_{t,k}^{-1}
\label{eq:1}
\end{equation}

where $t$ indexes the one-second integration step and $k \in [1, N]$ indexes samples within a one-second window. In the $n$-frame the $z$-axis points opposite to gravity, $\mathbf{g} = [0, 0, -g]^{\top}$; gravity alignment decouples the vertical motion features from the horizontal ones. The horizontal plane contains no absolute reference vector and the yaw angle drifts slowly, so the system operates with a frame that is not absolutely oriented and with a slowly evolving heading.

At time step $t$, the model receives the local inertial sequence $\mathbf{X}_{t}^{I} = \{[\boldsymbol{\omega}_{t,k}^{n}, \mathbf{f}_{t,k}^{n}]\}_{k=1}^{N} \in \mathbb{R}^{N \times 6}$ of the current 1 s window and outputs the 3D relative displacement increment over that interval, $\Delta\mathbf{p}_{t} = [\Delta x_{t}, \Delta y_{t}, \Delta z_{t}]^{\top}$. The global position is obtained by discrete accumulation:

\begin{equation}
\mathbf{p}_{t} = \mathbf{p}_{t-1} + \Delta\mathbf{p}_{t}
\label{eq:2}
\end{equation}

The barometric branch of the complete architecture is pluggable. When a valid barometric reading exists at inference, it is admitted as an observation input; when it is missing, the branch is switched off by a mask and the same weights fall back to a pure inertial pathway. The implementation of both modes is described in Section~\ref{sec:3.4}.

\subsection{Under-Excited Motion and Local Observability Analysis}
\label{sec:3.2}

The continuous kinematics of the pedestrian in the local gravity-aligned frame satisfies

\begin{equation}
\dot{\mathbf{p}}(t) = \mathbf{v}(t), \qquad \dot{\mathbf{v}}(t) = \mathbf{f}^{n}(t) + \mathbf{g}
\label{eq:3}
\end{equation}

where $\mathbf{f}^{n}(t)$ is the specific force vector and $\mathbf{g}$ is the local gravity. During level walking, the alternating collisions between the foot and the ground excite a periodic 1.5--2.5 Hz impact on the accelerometer, which supplies a biomechanical prior for instantaneous velocity estimation; a 1 s window suffices to distinguish common walking gaits.

In scenes of mechanical transport such as vertical elevators and escalators, this premise no longer holds. Taking a vertical elevator as an example, the car accelerates for 1--3 s and then cruises at constant velocity for ten to tens of seconds; the net motion acceleration vanishes ($a_{\text{kin}} \equiv 0$) and the specific force degenerates into the reaction of static gravity:

\begin{equation}
\mathbf{f}^{n}(t) = -\mathbf{g} = [0, 0, g]^{\top}
\label{eq:4}
\end{equation}

Inertial measurements carry no velocity term in any regime. A window-based estimator infers velocity only through the statistical association between gait-induced signal structure and speed that it has learned, so its estimate is informative exactly when the distribution of the window depends on velocity. At the onset of a constant-velocity cruise, foot impact disappears and the window content becomes indistinguishable from that of a static window, independently of the velocity carried into the cruise:

\begin{equation}
p(\mathbf{X}_{t}^{I}\mid \mathbf{v}_{t}) \approx p_{\text{static}}(\mathbf{X}_{t}^{I}),
\qquad
I(\mathbf{X}_{t}^{I};\mathbf{v}_{t}) \approx 0
\label{eq:5}
\end{equation}

where $p_{\text{static}}$ is the distribution of a window recorded while standing still and $I(\cdot;\cdot)$ denotes mutual information. The observation no longer carries information about velocity, so the posterior of any window-based estimator reduces to its prior; a short-window network trained with the negative log-likelihood of Section~\ref{sec:3.5.1} then returns the conditional mean of the training set, and the vertical displacement collapses toward zero.

This unobservability is with respect to the window-based observation model, not with respect to the dynamics of the pedestrian: what the short-window model lacks is a mechanism that carries the state across the cruise and is triggered at the onset event. The viable route is therefore to establish causal state continuity across the excitation-free gap, so that a long-range causal context latches the transient acceleration impulse at the moment of entering the transport segment and maintains the motion state throughout the cruise. Section~\ref{sec:3.3} designs LTCA accordingly, and Section~\ref{sec:4.3} examines this mechanism through history truncation and attention intervention.

\subsection{LTCA Architecture}
\label{sec:3.3}

Computing global self-attention over tens of seconds of raw 100 Hz sequences costs quadratically in the number of samples, and high-frequency gait impacts also disturb the attention allocation. LTCA therefore adopts a hierarchical design that combines bottom-level local aggregation with top-level bounded causal attention.

\subsubsection{Bottom-Level Local Motion Token Self-Attention Pooling}
\label{sec:3.3.1}

The receptive field of a feed-forward convolutional network is bounded, the local estimation residuals of different windows are approximately zero-mean and independent, and they cancel when the displacements are summed \cite{b4}. The bottom-level backbone inherits this property: a one-dimensional residual convolutional network takes the 1 s local inertial sequence as input and produces dense features $\mathbf{H}_{t} \in \mathbb{R}^{N \times C}$ with $N = 100$.

Within a single gait cycle, the energy of heel strike, stance, and push-off is unevenly distributed over the contact phase, and global average pooling with fixed uniform weights smooths away the transient impact features. We therefore introduce self-attention pooling with a learnable token: while preserving residual cancellation, it compresses the 1 s dense features into a single token that represents the current motion state. The local features are first linearly projected and augmented with a one-dimensional positional encoding:

\begin{equation}
\mathbf{M}_{t} = \text{LayerNorm}(\mathbf{H}_{t}\mathbf{W}_{\text{proj}} + \mathbf{b}_{\text{proj}}) + \mathbf{P}_{\text{local}}
\label{eq:6}
\end{equation}

Following the class-token aggregation of vision Transformers \cite{b18}, a learnable token vector $\mathbf{e}_{\text{token}} \in \mathbb{R}^{1 \times d}$ is prepended to the feature sequence, forming the augmented feature matrix $\mathbf{Z}_{t}^{0} = [\mathbf{e}_{\text{token}}; \mathbf{M}_{t}] \in \mathbb{R}^{(N+1) \times d}$, which is then passed through one layer of multi-head self-attention and a feed-forward network:

\begin{equation}
\mathbf{Z}_{t}^{1} = \text{MHSA}(\text{LayerNorm}(\mathbf{Z}_{t}^{0})) + \mathbf{Z}_{t}^{0}
\label{eq:7}
\end{equation}

\begin{equation}
\mathbf{Z}_{t}^{2} = \text{FFN}(\text{LayerNorm}(\mathbf{Z}_{t}^{1})) + \mathbf{Z}_{t}^{1}
\label{eq:8}
\end{equation}

The hidden dimension is $d = 128$. The output vector at the first position of the augmented sequence, $\mathbf{u}_{t} = \mathbf{Z}_{t}^{2}[0, :] \in \mathbb{R}^{d}$, represents the kinematic context of the current 1 s window and serves as a compact motion token for the top-level long-range attention.

\subsubsection{Top-Level Bounded Causal Temporal Attention (BCTA)}
\label{sec:3.3.2}

At a discrete integration step $t$, the top-level bounded causal temporal attention (BCTA) receives the sequence of $L = 60$ historical tokens that includes the current step:

\begin{equation}
\mathbf{S}_{t} = [\mathbf{u}_{t-L+1}, \ldots, \mathbf{u}_{t}]^{\top} \in \mathbb{R}^{L \times d}
\label{eq:9}
\end{equation}

The top-level attention builds causal associations between the current estimate and preceding motion events; the fixed-length sliding window bounds the effective memory range and constitutes a finite-impulse-response mapping, so that distant history perturbations do not propagate without bound.

The top level stacks two layers of multi-head attention and introduces a one-dimensional learnable relative temporal bias $b^{\ell}(t-j)$ \cite{b19} into the attention score matrix for $0 \leq t-j < L$, explicitly reflecting the temporal distance between a historical step and the current one:

\begin{equation}
A^{\ell}(t,j) = \frac{(\mathbf{u}_{t}\mathbf{W}_{Q}^{\ell})(\mathbf{u}_{j}\mathbf{W}_{K}^{\ell})^{\top}}{\sqrt{d_{k}}} + b^{\ell}(t-j)
\label{eq:10}
\end{equation}

Together with a strict lower-triangular causal mask, the output of a single head is

\begin{equation}
\begin{aligned}
\alpha^{\ell}(t,j) &= \frac{\exp(A^{\ell}(t,j))}{\sum_{\tau=t-L+1}^{t}\exp(A^{\ell}(t,\tau))},\\
\mathbf{o}_{t}^{\ell} &= \sum_{j=t-L+1}^{t}\alpha^{\ell}(t,j)(\mathbf{u}_{j}\mathbf{W}_{V}^{\ell})
\end{aligned}
\label{eq:11}
\end{equation}

The outputs of all heads are concatenated, linearly projected, and added through a residual connection:

\begin{equation}
\mathbf{s}_{t} = \text{LayerNorm}\left(\mathbf{u}_{t} + [\mathbf{o}_{t}^{1}, \ldots, \mathbf{o}_{t}^{H}]\mathbf{W}_{O}\right)
\label{eq:12}
\end{equation}

\begin{equation}
\mathbf{z}_{t} = \text{LayerNorm}\left(\mathbf{s}_{t} + \text{FFN}(\mathbf{s}_{t})\right)
\label{eq:13}
\end{equation}

The feature vector $\mathbf{z}_{t}$ is fed to a linear prediction head that outputs the mean and the per-axis log-standard deviation of the 1 s displacement. The mean is used for displacement estimation and the log-standard deviation enters the heteroscedastic loss of Section~\ref{sec:3.5.1}; this design follows native TLIO.

\subsubsection{Online Streaming Inference and Key-Value Caching}
\label{sec:3.3.3}

Online inference maintains a first-in-first-out key-value cache of length $L = 60$ in memory. Whenever a new 1 s measurement arrives, the bottom level computes only the query and the key-value pair of the current token; the cache pops the oldest key-value pair and pushes the newest one, and the current query attends to the historical keys. The per-step cost of this design is independent of the total trajectory duration.

In the evaluation implementation, accuracy is measured by recomputing, at each step, the 60 cached tokens together with the current token (61 one-second windows in total) rather than by carrying the cache across steps, whereas the latency evaluation of Section~\ref{sec:4.6} additionally reports the cost with an explicit cache. Both use the same weights and produce identical per-token outputs. The cache removes the cost of recomputing historical tokens, so the latency reported in Section~\ref{sec:4.6} represents what deployment can achieve.

\subsection{Pluggable Local-Differential Barometric Fusion and Paired Conditional Training}
\label{sec:3.4}

\subsubsection{Local Differential Features}
\label{sec:3.4.1}

A barometer outputs absolute static pressure, whose macroscopic component varies with floor and weather, so direct input would mix the floor height into the motion features. Pressure is also affected by slow weather changes and airflow disturbances, and may be interrupted by sleep or communication anomalies. The feature design must therefore be simultaneously invariant to floor translation, markable when missing, and causal. Let $h_{t,k}$ be the height converted from static pressure at the $k$-th sample of the $t$-th 1 s window; we construct a local height-difference feature sequence $\mathbf{F}_{t}^{\text{baro}} = \{[\Delta h_{t,k}, m_{t,k}^{B}, \delta t_{t,k}^{B}]\}_{k=1}^{N} \in \mathbb{R}^{N \times 3}$, whose three components are:

\begin{enumerate}
\item the relative height difference $\Delta h_{t,k} = h_{t,k} - h_{t,1}$, referenced to the first valid sample of the current window. This difference cancels the macroscopic absolute-altitude offset of different floors and brings the relative height difference to the same order of magnitude as velocity;
\item the sampling-validity binary mask $m_{t,k}^{B} \in \{0, 1\}$, which marks the arrival instants of valid readings and lets the network distinguish ``a reading of zero'' from ``no reading'';
\item the sampling-interval normalization $\delta t_{t,k}^{B} \in [0, 1]$, which records the relative time since the last valid sample and lets the network judge the freshness of the current reading.
\end{enumerate}

These three features use only the current window and preceding samples, introduce no future information, and can be computed window by window. The window-relative difference removes the floor translation that is irrelevant to motion, so the network only has to learn the height change within the window. The sequence is mapped to a low-dimensional feature $\mathbf{G}_{t}^{B} \in \mathbb{R}^{16}$ by a one-dimensional causal convolution and a smooth GELU network, and enters the fusion module as the barometric branch.

\subsubsection{Identity Pass-Through Projection and Hard Gating}
\label{sec:3.4.2}

Fusion adopts an identity pass-through projection, so that the forward pass relies on no imputation when the barometer is missing:

\begin{equation}
\mathbf{H}_{t}^{\text{fused}} = \mathbf{W}_{\text{proj}}\left[\mathbf{H}_{t}^{\text{IMU}} ; \mathbf{G}_{t}^{B} \cdot m_{t}^{\text{win}} ; c_{t}^{\text{win}} \cdot m_{t}^{\text{win}}\right]
\label{eq:14}
\end{equation}

where $\mathbf{H}_{t}^{\text{IMU}}$ denotes the shallow features of the inertial backbone, $m_{t}^{\text{win}}$ the window-level barometer availability flag, and $c_{t}^{\text{win}}$ the coverage ratio of valid measurements within the window. The projection matrix is initialized to the identity on the inertial block and to a small-variance Gaussian perturbation on the barometric and coverage blocks, which breaks symmetry and yields non-zero gradients.

The fusion gate is a hard gate. When $m_{t}^{\text{win}} = 0$, the barometric and coverage blocks in Eq.~(\ref{eq:14}) are set exactly to zero, and the output reduces to the inertial block acting on $\mathbf{H}_{t}^{\text{IMU}}$, without interpolation, extrapolation, or missing-value estimation. When only part of the samples in a window are valid, $m_{t}^{\text{win}}$ may still be set to 1, and the sparsity is characterized by the coverage ratio.

The inertial block of the identity pass-through is initialized to the identity matrix so that fusion starts from an approximately identity mapping; after training the block is no longer constrained to the identity, and the statement therefore describes the initialization and the design intent. Whether the system is online at inference is decided solely by $m_{t}^{\text{win}}$, independently of the network weights.

\subsubsection{Structural and Training Layers of the Pluggable Strategy}
\label{sec:3.4.3}

Identity pass-through and hard gating guarantee only that switching is computationally feasible, not that the accuracy after switching is acceptable. The pluggable strategy is therefore split into a structural layer and a training layer.

The structural layer decides switching at inference. $m_{t}^{\text{win}} = 1$ takes the fused forward pass, and $m_{t}^{\text{win}} = 0$ takes the fallback forward pass. The two passes share one set of weights, require no retraining, and the fallback path is computationally equivalent to the pure inertial forward pass.

The training layer decides whether the fallback path retains the gain. Each sample is forwarded through both the online and the fallback paths, and the endpoint negative log-likelihood (NLL) of the two paths is optimized jointly, with both losses sharing the same backbone and projection matrix.

Paired training also brings a training-time property. In the online mode, the barometer is an inference-time input and the setting is ordinary multimodal fusion; in the fallback mode, the barometer is unavailable at inference and participates only through the online branch during training, where it shapes the shared representation and thus acts as a privileged observation in the sense of LUPI. Because the two paths share all parameters and the same labels, the joint objective is equivalent to a single risk evaluated over the mixture of the two input conditions; we therefore use LUPI as an interpretive lens rather than as an implementation of a teacher--student scheme.

This is structurally close to generalized distillation \cite{b14}. The online branch provides privileged supervision, the masked branch is the student path executed at deployment, and the two share parameters. Pluggability is therefore the combined effect of a structural hard switch and training-time privileged shaping.

\subsection{Training Objective and Training Setup}
\label{sec:3.5}

\subsubsection{Training Objective}
\label{sec:3.5.1}

Instead of the native whole-window position loss, LTCA is supervised by the Gaussian negative log-likelihood on the 1 s endpoints. At a supervised endpoint $t$, the network outputs the displacement mean $\widehat{\Delta\mathbf{p}}_{t} \in \mathbb{R}^{3}$ and the log-standard deviation $\widehat{\mathbf{s}}_{t} \in \mathbb{R}^{3}$ of the 1 s interval, with ground truth $\Delta\mathbf{p}_{t}$; writing $\widehat{\sigma}_{t,i} = \exp(\widehat{s}_{t,i})$, the single-endpoint loss is

\begin{equation}
\mathcal{L}_{t} = \frac{1}{2}\sum_{i \in \{x,y,z\}}\left[\frac{(\Delta p_{t,i} - \widehat{\Delta p}_{t,i})^{2}}{\widehat{\sigma}_{t,i}^{2}} + 2\widehat{s}_{t,i}\right]
\label{eq:15}
\end{equation}

The batch loss is the mean of $\mathcal{L}_{t}$ over all supervised endpoints in the batch, and for 2D tasks the sum runs over the two horizontal axes only. This form follows native TLIO and lets the network estimate the confidence of each one-second displacement. Models with the barometric branch compute the online and fallback losses separately and back-propagate them jointly.

Variance warm-up is used in the early stage of training: for the first $N_{w}$ epochs the standard deviation is fixed to 1 and only the displacement mean is updated; once the mean has stabilized, the log-standard deviation is released and the two terms are optimized jointly.

\subsubsection{Sample Organization and Supervision Protocol}
\label{sec:3.5.2}

A training sample consists of 61 consecutive 1 s windows. The first 60 serve as historical context and the last one as the supervised endpoint, and the loss is computed only at that endpoint: one forward pass reads 60 s of history and outputs only the displacement of the current 1 s.

The training set is traversed in rotation over 10 phase offsets, that is, Phase-10 sampling. The same trajectory is cut at different time origins in different epochs, so that the same motion appears repeatedly under different historical phases and the bias of block-wise and phase alignment is reduced.

To match this sample structure, the block-level augmentation parameters, including random yaw, gravity tilt, and constant bias, are sampled once for the whole 61-window sample and kept fixed. Per-window sampling would break the unified coordinate frame within the long history; this convention is also the premise of the per-window perturbation in Section~\ref{sec:3.5.3}.

\subsubsection{Data Augmentation}
\label{sec:3.5.3}

Data augmentation falls into inherited and new items. The inherited items follow the native TLIO configuration and include random yaw in the horizontal plane, a $5^{\circ}$ gravity tilt, and constant biases of 0.2 and 0.05. Random yaw and gravity tilt act on the input and the displacement label through a single rigid rotation, and the constant bias is added to the sensor readings. Both backbones enable this augmentation set on the 3D MSIO task.

The new item is an independent per-step rotation perturbation, hereafter referred to as the rotation perturbation. On top of the unified coordinates provided by the block-level augmentation, we sample, independently for each 1 s window of a sample, a small-angle rotation with zero mean and standard deviation $\sigma = 1^{\circ}$, and apply it synchronously to the $n$-frame input and the displacement label of that window. In 3D the rotation is parameterized as a vector in $\mathfrak{so}(3)$; in 2D it degenerates to a yaw rotation about the vertical axis.

The rotation perturbation targets cross-step coordinate consistency rather than single-step displacement accuracy. Block-level augmentation provides a unified coordinate frame for the whole 60 s history, and the long-horizon attention can lower the training loss by exploiting the consistency of absolute orientation across steps; in open-loop evaluation and under real attitude drift, however, the coordinate frames of adjacent steps do not coincide, and this shortcut fails at inference. Per-window rotation makes the orientations of adjacent windows mutually inconsistent, so the model can only rely on the within-window relative geometry and rotation-invariant features.

The remaining training configuration follows the native implementation.

\section{Experiments}
\label{sec:4}

\subsection{Experimental Setup}
\label{sec:4.1}

\subsubsection{Dataset}
\label{sec:4.1.1}

Evaluation is conducted on our MSIO benchmark and on the public RoNIN and TLIO benchmarks. MSIO (Multi-Story Indoor Odometry) is a 3D localization benchmark for multistory buildings, collected in nine buildings that cover teaching buildings, laboratory buildings, and large commercial complexes, with motion modes including flat walking, stairs, escalators, and elevators. Data were collected with handheld devices; an external 100 Hz six-axis micro-electromechanical (MEMS) IMU and a barometric module recorded synchronously, and the reference trajectory was obtained with Odin1, a LiDAR-inertial mapping device.

MSIO is partitioned into four mutually exclusive parts (Table~\ref{tab:1}). The training set contains 290 sequences totaling 29.34 h; the validation set contains 30 sequences totaling 2.64 h; the Seen test set contains 30 sequences totaling 2.61 h and covers buildings seen during training; the Unseen test set contains 17 sequences totaling 1.67 h, recorded by test subjects and in buildings that do not appear in training. The Unseen split therefore measures generalization to both an unseen environment and an unseen subject pool. The validation set is used only for model selection, and neither test set participates in training or model selection.

\begin{table}[!t]
\centering
\caption{Splits of the MSIO dataset. All sequences were recorded with a 100 Hz six-axis IMU and a synchronized barometer; reference trajectories were generated by LiDAR-inertial mapping. The Unseen split participates in neither training nor model selection.}
\label{tab:1}
\footnotesize
\begin{tabular}{@{}lrcp{1.55in}@{}}
\toprule
Split & Seq. & Dur. (h) & Description\\
\midrule
Training & 290 & 29.34 & Used for training \\
Validation & 30 & 2.64 & Used for model selection \\
Seen test & 30 & 2.61 & Buildings seen in training; sessions disjoint from training \\
Unseen test & 17 & 1.67 & Buildings and test subjects unseen in training \\
\bottomrule
\end{tabular}
\end{table}

MSIO covers 3D multistory scenes, and inter-floor transit involves two passive transport modes, elevators and escalators. RoNIN \cite{b4} provides 200 Hz inertial data with ground truth from a Google Tango visual-inertial system; we evaluate planar dead reckoning on this benchmark under the attitude estimated onboard the phone. TLIO \cite{b5} provides 200 Hz head-mounted data, and its test set contains 36 long sequences with attitude obtained by open-loop integration of the onboard gyroscope; we use it to examine dead reckoning under attitude drift.

\subsubsection{Evaluation Metrics}
\label{sec:4.1.2}

All evaluations adopt open-loop dead reckoning. Each model runs with its native window and stride, integrates continuously from the start of the sequence, and is evaluated under a unified 0 s cold start, with no intermediate reset and no global pose registration. The per-mode breakdown of Table~\ref{tab:3} is diagnostic: it re-anchors each segment at its own start so that the error inside a mode is not dominated by drift accumulated in earlier segments, and it is therefore not comparable with Table~\ref{tab:2}.

\begin{itemize}
\item \textbf{Absolute trajectory error (ATE)} \cite{b20}: the root-mean-square position error, reported in three statistics, namely the 3D total error (ATE-3D), the horizontal-plane error (ATE-XY), and the vertical elevation error (Z-RMSE).
\item \textbf{Time-based relative trajectory error (RTE-60s)}: the root-mean-square relative displacement error over a fixed 60 s time span, computed by sliding over the whole sequence.
\item \textbf{Distance-based relative trajectory error (DTE-1m)}: the root-mean-square relative displacement error over a fixed 1 m traveled-distance span, computed by sliding over the whole sequence.
\end{itemize}

\subsubsection{Baselines}
\label{sec:4.1.3}

The baselines fall into three groups. The short-window baselines include Native RoNIN, Native TLIO, CTIN, and EqNIO-O(2); Native RoNIN uses its native 2 s window and the others use a 1 s window. The external long-horizon baseline is IDOL, which uses a bidirectional recurrent network with a training window expanded to 20 s by curriculum learning and, following its official deployment, is inferred with a 1 s window, since its bidirectional 20 s window needs future samples and cannot be streamed causally (see the note to Table~\ref{tab:2}). The third group consists of component controls of the complete architecture.

The complete architecture is instantiated on both backbones. The pure inertial instances are RoNIN-LTCA and TLIO-LTCA; the instances with the barometric branch are RoNIN-LTCA+Baro and TLIO-LTCA+Baro, each providing an online and a fallback mode. To isolate the contribution of the barometric observation itself, we additionally connect both native backbones to the same fusion module, yielding Native RoNIN+Baro and Native TLIO+Baro.

\subsubsection{Implementation Details}
\label{sec:4.1.4}

The training setup follows the native implementations, and only the parts that differ from them or are introduced here are described. TLIO is trained with Adam at a learning rate of $1 \times 10^{-4}$ and RoNIN with Adam at $3 \times 10^{-4}$; the effective batch size of LTCA is 128 for TLIO and 64 for RoNIN, with mixed precision on a single GPU.

\subsection{Evaluation on MSIO}
\label{sec:4.2}

We first compare the complete architecture with all existing methods, and then examine the accuracy of the same weights with the barometer cut off. Table~\ref{tab:2} reports ATE-3D together with its horizontal component ATE-XY and its vertical component Z-RMSE, as well as RTE-60s and DTE-1m, for the Seen and Unseen splits. The configuration columns give the model and the inference mode.

\begin{table*}[!t]
\centering
\caption{3D dead reckoning results on MSIO. For each backbone, ``online'' and ``fallback'' are two inference modes of the same set of weights; IDOL is inferred with a 1 s window following its official deployment configuration. Best results on each split are in bold. Unit: m.}
\label{tab:2}
\footnotesize
\begin{tabular}{@{}lllcccccc@{}}
\toprule
\multicolumn{9}{@{}l}{\textbf{(a) Seen}}\\
Model & LTCA & Baro & Mode & ATE-3D & ATE-XY & Z-RMSE & RTE-60s & DTE-1m \\
\midrule
Native RoNIN & $\times$ & $\times$ & -- & 5.64 & 3.93 & 3.54 & 3.60 & 0.31 \\
Native TLIO & $\times$ & $\times$ & -- & 5.91 & 3.57 & 4.23 & 4.01 & 0.31 \\
IDOL & $\times$ & $\times$ & -- & 5.87 & 3.95 & 3.89 & 3.99 & 0.32 \\
CTIN & $\times$ & $\times$ & -- & 5.68 & 3.52 & 3.90 & 3.69 & 0.30 \\
EqNIO & $\times$ & $\times$ & -- & 5.48 & 3.45 & 3.80 & 3.67 & 0.30 \\
RoNIN-LTCA & $\checkmark$ & $\times$ & -- & 3.44 & 3.26 & 0.86 & 2.02 & 0.18 \\
TLIO-LTCA & $\checkmark$ & $\times$ & -- & 3.26 & 2.85 & 1.18 & 1.96 & 0.19 \\
Native RoNIN+Baro & $\times$ & $\checkmark$ & online & 3.49 & 3.26 & 1.03 & 2.30 & 0.30 \\
Native RoNIN+Baro & $\times$ & $\times$ & fallback & 5.21 & 3.46 & 3.43 & 3.51 & 0.34 \\
Native TLIO+Baro & $\times$ & $\checkmark$ & online & 4.00 & 3.45 & 1.47 & 2.58 & 0.25 \\
Native TLIO+Baro & $\times$ & $\times$ & fallback & 5.79 & 3.49 & 4.14 & 3.85 & 0.31 \\
RoNIN-LTCA+Baro & $\checkmark$ & $\checkmark$ & online & 2.83 & 2.67 & 0.76 & 1.84 & 0.17 \\
RoNIN-LTCA+Baro & $\checkmark$ & $\times$ & fallback & 3.24 & 2.70 & 1.30 & 2.14 & 0.19 \\
TLIO-LTCA+Baro & $\checkmark$ & $\checkmark$ & online & \textbf{2.44} & 2.32 & 0.64 & \textbf{1.65} & \textbf{0.16} \\
TLIO-LTCA+Baro & $\checkmark$ & $\times$ & fallback & 2.64 & 2.34 & 0.95 & 1.84 & 0.17 \\
\midrule
\multicolumn{9}{@{}l}{\textbf{(b) Unseen}}\\
Model & LTCA & Baro & Mode & ATE-3D & ATE-XY & Z-RMSE & RTE-60s & DTE-1m \\
\midrule
Native RoNIN & $\times$ & $\times$ & -- & 7.80 & 5.78 & 4.47 & 4.28 & 0.49 \\
Native TLIO & $\times$ & $\times$ & -- & 8.10 & 5.76 & 5.12 & 4.69 & 0.48 \\
IDOL & $\times$ & $\times$ & -- & 7.92 & 4.98 & 5.52 & 4.87 & 0.48 \\
CTIN & $\times$ & $\times$ & -- & 7.61 & 5.56 & 4.40 & 4.30 & 0.54 \\
EqNIO & $\times$ & $\times$ & -- & 8.08 & 5.77 & 4.98 & 4.58 & 0.48 \\
RoNIN-LTCA & $\checkmark$ & $\times$ & -- & 4.55 & 3.98 & 1.79 & 2.51 & 0.25 \\
TLIO-LTCA & $\checkmark$ & $\times$ & -- & 6.91 & 5.70 & 2.98 & 3.50 & 0.48 \\
Native RoNIN+Baro & $\times$ & $\checkmark$ & online & 5.51 & 5.19 & 1.56 & 2.88 & 0.42 \\
Native RoNIN+Baro & $\times$ & $\times$ & fallback & 7.05 & 5.18 & 4.03 & 4.02 & 0.52 \\
Native TLIO+Baro & $\times$ & $\checkmark$ & online & 5.91 & 5.63 & 1.54 & 3.20 & 0.46 \\
Native TLIO+Baro & $\times$ & $\times$ & fallback & 8.13 & 5.72 & 5.25 & 4.73 & 0.54 \\
RoNIN-LTCA+Baro & $\checkmark$ & $\checkmark$ & online & \textbf{4.18} & 3.94 & 1.07 & \textbf{2.30} & \textbf{0.23} \\
RoNIN-LTCA+Baro & $\checkmark$ & $\times$ & fallback & 4.28 & 3.78 & 1.42 & 2.45 & 0.24 \\
TLIO-LTCA+Baro & $\checkmark$ & $\checkmark$ & online & 5.12 & 4.61 & 1.56 & 2.78 & 0.24 \\
TLIO-LTCA+Baro & $\checkmark$ & $\times$ & fallback & 5.23 & 4.51 & 1.84 & 2.87 & 0.25 \\
\bottomrule
\end{tabular}
\end{table*}

\begin{figure*}[!t]
\centering
\includegraphics[width=\textwidth]{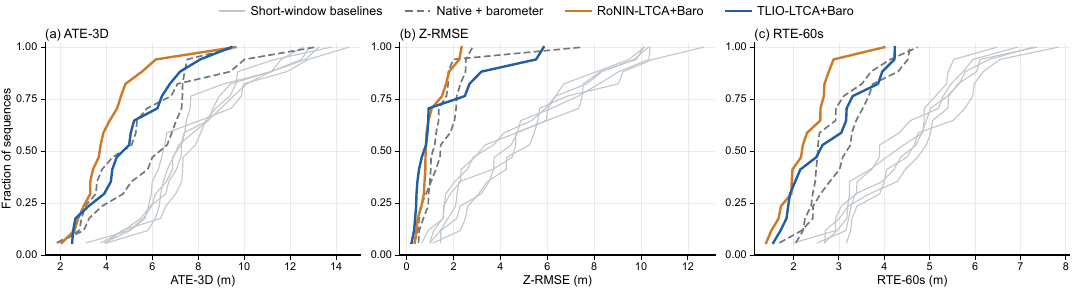}
\caption{Error cumulative distributions over the 17 Unseen sequences: (a) ATE-3D; (b) Z-RMSE; (c) RTE-60s ($n=17$).}
\label{fig:1}
\end{figure*}

\begin{figure*}[!t]
\centering
\includegraphics[width=\textwidth]{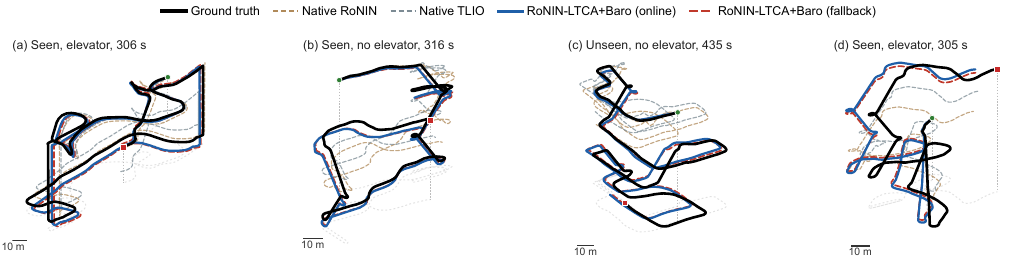}
\caption{Planar trajectories of four representative sequences: (a) and (d) are Seen sequences that contain elevator segments, (b) is a Seen sequence without elevators, and (c) is an Unseen sequence. Ground truth, the native baselines, and the complete architecture in online and fallback modes are compared.}
\label{fig:2}
\end{figure*}

In the following, RoNIN-LTCA+Baro and TLIO-LTCA+Baro are abbreviated as the RoNIN instance and the TLIO instance.

Under online inference, both LTCA instances outperform every comparison in the table. The TLIO instance attains an ATE-3D of 2.44 m on Seen and 5.12 m on Unseen, and the RoNIN instance attains 2.83 m and 4.18 m on the same two splits. Both remain below every short-window baseline, namely 5.48--5.91 m on Seen and 7.61--8.10 m on Unseen, and also below the native+barometer models at 3.49--4.00 m and 5.51--5.91 m.

RTE-60s and DTE-1m are 1.65--1.84 m and 0.16--0.17 m on Seen, and 2.30--2.78 m and 0.23--0.24 m on Unseen, all lower than the non-LTCA comparisons, which shows that the improvement is not confined to endpoint error. IDOL, evaluated with its 1 s deployment window, stays within the error range of the non-LTCA comparisons. Figure~\ref{fig:1} shows the corresponding error distributions over the 17 Unseen sequences, and Fig.~\ref{fig:2} compares the planar trajectories of four representative sequences against the native baselines and the complete architecture.

After the barometer is cut off, the same weights of the TLIO instance fall back to 2.64/5.23 m and those of the RoNIN instance to 3.24/4.28 m, still below every short-window baseline and every native+barometer model.

Even the model trained without any barometric input reaches a high accuracy on elevator segments. For RoNIN it matches the barometer-assisted instance (Z-RMSE 0.43 versus 0.39 m on Seen and 0.61 versus 0.47 m on Unseen), whereas for TLIO the barometer still roughly halves the vertical error (0.63 versus 0.27 m on Seen). The long-horizon structure is therefore effective for elevator estimation on its own, and the barometer adds a further, backbone-dependent margin.

\begin{table*}[!t]
\centering
\caption{Diagnostic error breakdown by motion mode; each cell reports ATE-XY / Z-RMSE. This table is not comparable with Table II, because within each segment the origin is reset to the segment start and the RMS is computed per segment, then averaged over segments of the same mode; ``online'' and ``fallback'' are two inference modes of the same weights. Unit: m.}
\label{tab:3}
\footnotesize
\begin{tabular}{@{}lcccc@{}}
\toprule
\multicolumn{5}{@{}l}{\textbf{(a) Seen}}\\
Model & Flat & Stairs & Escalator & Elevator \\
\midrule
\textbf{Native RoNIN} & 1.10 / 0.18 & 0.54 / 0.16 & 1.23 / 1.82 & 0.28 / 2.79 \\
\textbf{Native TLIO} & 1.07 / \textbf{0.08} & 0.44 / 0.13 & 1.66 / 1.93 & 0.16 / 3.76 \\
\textbf{RoNIN-LTCA} & 0.92 / 0.12 & 0.49 / 0.11 & 0.75 / 0.36 & 0.14 / 0.43 \\
\textbf{TLIO-LTCA} & 0.91 / 0.12 & 0.42 / 0.11 & 0.65 / 0.34 & 0.14 / 0.63 \\
\textbf{RoNIN-LTCA+Baro (online)} & 0.84 / 0.12 & 0.46 / 0.10 & 0.67 / 0.36 & 0.14 / 0.39 \\
\textbf{RoNIN-LTCA+Baro (fallback)} & 0.84 / 0.14 & 0.47 / 0.11 & 0.67 / 0.67 & 0.14 / 0.67 \\
\textbf{TLIO-LTCA+Baro (online)} & \textbf{0.81} / 0.15 & 0.43 / 0.09 & 0.57 / 0.22 & 0.13 / 0.27 \\
\textbf{TLIO-LTCA+Baro (fallback)} & 0.82 / 0.15 & 0.43 / 0.09 & 0.56 / 0.48 & 0.13 / 0.40 \\
\midrule
\multicolumn{5}{@{}l}{\textbf{(b) Unseen}}\\
Model & Flat & Stairs & Escalator & Elevator \\
\midrule
\textbf{Native RoNIN} & 1.92 / 0.31 & 0.61 / 0.26 & 1.40 / 2.71 & 0.23 / 1.59 \\
\textbf{Native TLIO} & 1.95 / 0.22 & 0.53 / 0.30 & 1.59 / 2.79 & 0.18 / 2.32 \\
\textbf{RoNIN-LTCA} & 1.64 / 0.26 & 0.61 / 0.19 & 0.62 / 0.45 & 0.16 / 0.61 \\
\textbf{TLIO-LTCA} & 1.96 / 0.34 & 0.61 / 0.19 & 0.53 / 0.63 & 0.16 / 0.77 \\
\textbf{RoNIN-LTCA+Baro (online)} & 1.44 / 0.17 & 0.53 / 0.14 & 0.89 / 0.25 & 0.17 / 0.47 \\
\textbf{RoNIN-LTCA+Baro (fallback)} & 1.43 / 0.19 & 0.54 / 0.18 & 1.01 / 0.56 & 0.16 / 0.71 \\
\textbf{TLIO-LTCA+Baro (online)} & 1.83 / 0.22 & 0.43 / 0.13 & 0.83 / 0.52 & 0.13 / 0.50 \\
\textbf{TLIO-LTCA+Baro (fallback)} & 1.81 / 0.22 & 0.44 / 0.17 & 0.93 / 0.75 & 0.14 / 0.61 \\
\bottomrule
\end{tabular}
\end{table*}

Per-mode errors localize the gain to the transport segments (Table~\ref{tab:3}). For elevators, Z-RMSE in online mode is 0.27--0.39 m on Seen and 0.47--0.50 m on Unseen, whereas the short-window baselines give 2.79/3.76 m and 1.59/2.32 m. For escalators, ATE-XY drops from 1.23/1.66 m to 0.67/0.57 m and Z-RMSE from 1.82/1.93 m to 0.36/0.22 m. Flat walking and stairs change very little.

The barometer gives the instance models higher vertical accuracy, and even the fallback version estimates non-elevator segments such as flat walking and stairs better. This indicates that the barometer does not act only on elevation estimation but serves as a source of information that improves the model as a whole.

The overall error increase from Seen to Unseen comes mainly from horizontal drift during flat walking and over long sequences. In online mode, ATE-XY of the TLIO instance rises from 2.32 m to 4.61 m and that of the RoNIN instance from 2.67 m to 3.94 m; elevator segments do not become an error source.

\subsection{Mechanistic Analysis of LTCA}
\label{sec:4.3}

If the failure originates from an unobservable state, the short-window output should exhibit amplitude degradation rather than directional error. This is exactly what the MSIO elevator and escalator data show. The short-window model can still follow the ground truth during the acceleration phase, but degrades rapidly once constant-velocity cruise begins: the median vertical speed is 0.70 and 0.67 times the ground truth. The directional accuracy on amplitude-normal steps is 100\%, whereas on the degraded steps it is only 25.4\% and 37.0\% for elevators and escalators, respectively. What the model lacks is an amplitude observation.

\begin{figure*}[!t]
\centering
\includegraphics[width=\textwidth]{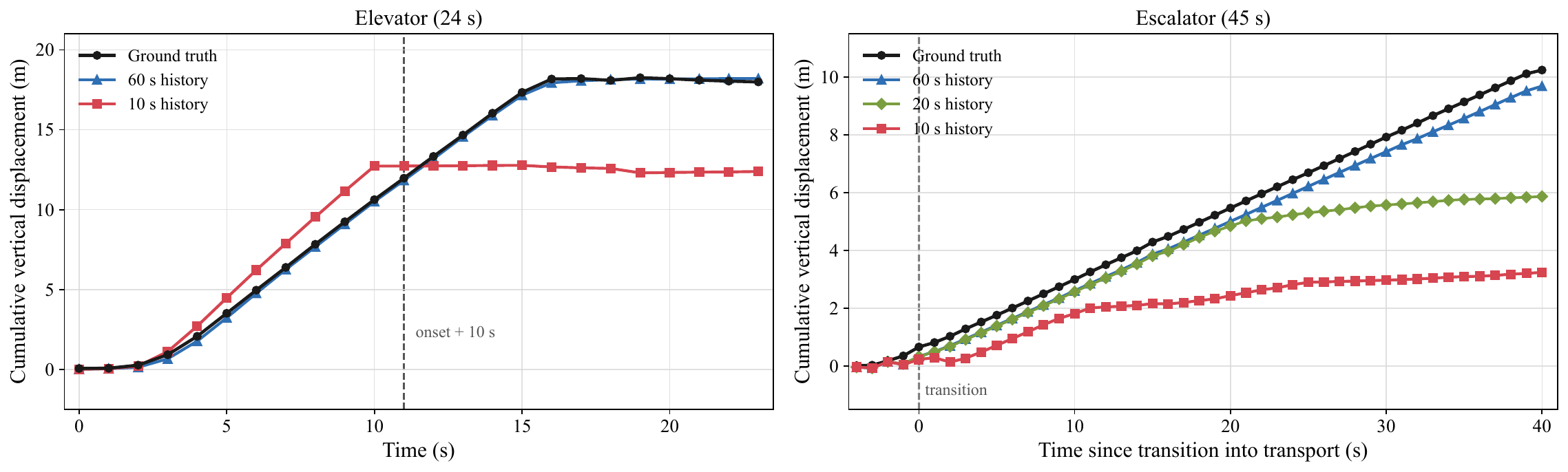}
\caption{Accumulated vertical displacement when the visible history is restricted: (a) 24 s elevator segment; (b) 45 s escalator segment. The horizontal axis is sequence time in (a) and the time since the onset event in (b).}
\label{fig:3}
\end{figure*}

If the state is maintained by history, restricting the visible history should make the state vanish as soon as the event slides out. Results obtained with the same 60 s weights and only a reduced visible history at inference support this, as shown in Fig.~\ref{fig:3}. In the 24 s elevator segment, a 10 s history stops accumulating once the event leaves the window, and the vertical displacement stalls at 12.4 m against a ground truth of 18.0 m, whereas the 60 s history accumulates to 18.2 m. In the 45 s escalator segment, the 10 s and 20 s histories stall at 3.2 m and 5.9 m, while the 60 s history reaches 9.7 m against a ground truth of 10.2 m.

Figure~\ref{fig:3}(a) shows the 24 s elevator segment and Fig.~\ref{fig:3}(b) the 45 s escalator segment. The short-history curves flatten after the vertical line, whereas the 60 s history keeps accumulating up to the ground truth. The three curves share one set of weights and differ only in the length of history visible at inference.

A stronger line of evidence comes from attention intervention, shown in Table~\ref{tab:4}. After masking attention to the tokens of the onset event, the vertical speed in the elevator cruise drops from 1.358 to 0.347 m/s, a reduction of 74\%, whereas masking control tokens gives 1.379 m/s. For escalators, masking the five tokens around the transition reduces the output from 0.242 to 0.175 m/s, a reduction of 28\%. Attention mass also concentrates near the event: the tokens of the elevator acceleration phase account for 3.9\% of the tokens but receive 15.7\% of the attention mass, and the escalator transport segment accounts for 20.9\% of the tokens but receives 37.9\%.

\begin{table}[!t]
\centering
\caption{Vertical speed output after masking attention to the onset-event tokens (60 s pure-IMU model, median). The model is the inertial-only model without the barometric branch; masking affects attention only and leaves all other inputs unchanged. Unit: m/s.}
\label{tab:4}
\footnotesize
\begin{tabular}{@{}lcc@{}}
\toprule
Input condition & Elevator & Escalator\\
\midrule
True input & 1.358 & 0.242 \\
Mask attention to event tokens & 0.347 ($-74\%$) & 0.175 ($-28\%$) \\
Mask attention to control tokens & 1.379 (+2\%) & 0.242 (0\%) \\
\bottomrule
\end{tabular}
\end{table}

Both interventions are single-variable operations. Truncation changes only the visible history, masking changes only the attention weights of the event tokens, and all other inputs stay unchanged; masking control tokens hardly changes the output.

The two transport modes draw their state from different sources but support the same mechanism. Elevator speed is determined by the impulse of the acceleration phase at the onset event, and the per-segment mean $|v_z|$ of different elevators ranges from 0.33 to 1.20 m/s. Escalator speed is set by the machine: over 43 segments the mean is 0.216 m/s with a between-segment standard deviation of 0.017, and the horizontal speed is 0.452 m/s; the transition phase only provides a criterion for whether the system is in transport and in which direction. Injecting vibration with an RMS of 0.1--0.7 m/s$^2$ into the current escalator window leaves the output at 0.24 m/s, whereas replacing the current window by the mean of vibration-free windows drops the output to 0.04 m/s. The amplitude evidently comes from the history.

The same mechanism also yields a requirement on window length. The visible history must cover the onset instant: when scanning with a 1 s stride, the median margin is 3 s for elevators and 2 s for escalators. If a single run outlasts the window, the state is lost again; in MSIO, elevator and escalator runs last no longer than 24 s and 45 s, which a 60 s window covers.

These results attribute the gain of LTCA to the latching and maintenance of the onset event by the causal history. Over long horizons the inertial channel does not gain new information; it maintains the state, and in doing so copes with under-excited motion states.

\subsection{Role of the Pluggable Barometer}
\label{sec:4.4}

The pluggable barometer can provide a direct elevation constraint. Table~\ref{tab:5} reports three configurations: A is a separately trained pure inertial model, B is paired conditional training with the barometer cut at inference, and C is paired conditional training with the barometer online at inference. The difference between B and C measures the margin of the direct elevation constraint, and the difference between A and B measures the additional benefit of the training-time condition.

\begin{table*}[!t]
\centering
\caption{Barometric branch and training conditions (TLIO backbone). A is the pure inertial control without the barometric branch and uses a single-path NLL; B and C are the two inference modes of paired conditional training, B being the fallback configuration in which the availability mask is set to zero at inference. For the RoNIN backbone, A is 3.44/4.55 m, B is 3.24/4.28 m, and C is 2.83/4.18 m. Unit: m.}
\label{tab:5}
\footnotesize
\begin{tabular}{@{}cp{1.55in}lcccc@{}}
\toprule
 & & Baro at & \multicolumn{2}{c}{ATE-3D} & \multicolumn{2}{c}{Z-RMSE}\\
\cmidrule(lr){4-5}\cmidrule(lr){6-7}
Cfg. & Training & inference & Seen & Unseen & Seen & Unseen\\
\midrule
A & single-path NLL (no barometric branch) & none & 3.26 & 6.91 & 1.18 & 2.98 \\
B & paired conditional training & cut (fallback) & 2.64 & 5.23 & 0.95 & 1.84 \\
C & paired conditional training & online & 2.44 & 5.12 & 0.64 & 1.56 \\
\bottomrule
\end{tabular}
\end{table*}

Configurations A, B, and C are evaluated on the same backbone, data splits, and 0 s cold-start protocol. B and C share one set of weights, and their difference stems only from whether the barometer is connected at inference. A and B differ in the branch setting and in the training scheme at the same time.

The value of the direct elevation constraint lies mainly in the vertical channel. Switching the same weights from fallback to online lowers the Z-RMSE of the TLIO instance from 0.95/1.84 m to 0.64/1.56 m on Seen and Unseen, whereas ATE-3D only drops from 2.64/5.23 m to 2.44/5.12 m, a difference of 0.20/0.11 m, because the overall error is dominated by horizontal drift. The RoNIN instance behaves in the same direction, with Z-RMSE dropping from 1.30 m to 0.76 m and ATE-3D from 3.24/4.28 m to 2.83/4.18 m.

The training-time condition brings an additional benefit. Configuration B performs fallback inference without any barometric input and yet achieves Seen and Unseen ATE-3D of 2.64/5.23 m, better than the separately trained pure inertial model A at 3.26/6.91 m, a reduction of 0.62 m and 1.68 m, respectively; Z-RMSE drops from 1.18/2.98 m to 0.95/1.84 m. The RoNIN instance moves in the same direction with a smaller magnitude, from A at 3.44/4.55 m to B at 3.24/4.28 m. This additional benefit is consistent with the privileged shaping expected from paired training.

The difference between A and B also includes the extra parameters of the branch and the auxiliary supervision of the two-path loss, so we state the correlation only and do not attribute the whole difference to the privileged-information mechanism.

\begin{figure}[!t]
\centering
\includegraphics[width=\columnwidth]{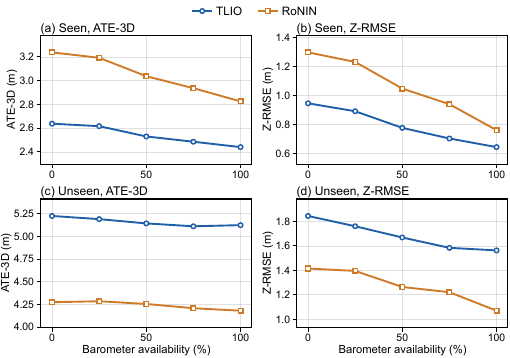}
\caption{Effect of the barometer availability ratio (0--100\%) on ATE-3D and Z-RMSE for both backbones: (a)(b) Seen; (c)(d) Unseen.}
\label{fig:4}
\end{figure}

Degradation is smooth when the barometer is unavailable (Fig.~\ref{fig:4}). As the availability ratio falls from 100\% to 0\%, Unseen ATE-3D of the two backbones rises from 5.12 m to 5.23 m and from 4.18 m to 4.28 m, about 2\%. Z-RMSE increases monotonically, from 1.56 m to 1.84 m and from 1.07 m to 1.42 m, respectively. At 75\% availability, Seen ATE-3D rises by 2\%--4\% and Z-RMSE by 9\%--24\%. The 0\% endpoint is the fallback configuration, consistent with configuration B in Table~\ref{tab:5}.

\subsection{Public Benchmarks and Ablations}
\label{sec:4.5}

\subsubsection{Results on Public Benchmarks}
\label{sec:4.5.1}

\begin{table}[!t]
\centering
\caption{Evaluation on the public RoNIN and TLIO benchmarks. Unit: m.}
\label{tab:6}
\scriptsize
\textbf{(a) RoNIN, 200 Hz, 2D (phone attitude)}\par\vspace{2pt}
\begin{tabular}{@{}llccc@{}}
\toprule
Model & Split & ATE & RTE-60s & DTE-1m\\
\midrule
Native RoNIN & Seen & 7.08 & 4.66 & 0.23 \\
Native RoNIN & Unseen & 10.48 & 7.29 & 0.33 \\
RoNIN-LTCA (60 s) & Seen & 6.14 & 4.06 & 0.18 \\
RoNIN-LTCA (60 s) & Unseen & 9.10 & 6.44 & 0.27 \\
\bottomrule
\end{tabular}
\vspace{6pt}

\textbf{(b) TLIO, 200 Hz, 3D (open-loop attitude, 36 sequences)}\par\vspace{2pt}
\begin{tabular}{@{}lccccc@{}}
\toprule
Model & ATE-3D & ATE-XY & Z-RMSE & RTE-60s & DTE-1m\\
\midrule
Native TLIO & 2.42 & 2.38 & 0.39 & 1.80 & 0.24 \\
TLIO-LTCA (60 s) & 2.14 & 2.08 & 0.40 & 1.68 & 0.22 \\
\bottomrule
\end{tabular}
\end{table}

With 60 s of causal history, the Seen/Unseen ATE of RoNIN drops from 7.08/10.48 m to 6.14/9.10 m, and the ATE-3D of TLIO drops from 2.42 m to 2.14 m while Z-RMSE stays at 0.39/0.40 m. RTE-60s and DTE-1m decrease as well. These public-benchmark numbers combine the long-horizon structure with the rotation augmentation of Section~\ref{sec:3.5.3}, and Table~\ref{tab:7} shows that the two are not separable here, since without the augmentation the long-horizon model is worse than its native counterpart. The gain is concentrated in the horizontal channel, as expected on these datasets, which contain no sustained elevator or escalator cruise across which a vertical state would have to be carried.

\subsubsection{Ablation on Data Augmentation}
\label{sec:4.5.2}

Among the data augmentation items, only the independent per-step rotation perturbation is introduced here; the others are inherited from the native TLIO configuration item by item.

\begin{table}[!t]
\centering
\caption{Ablation on data augmentation. Unit: m.}
\label{tab:7}
\footnotesize
\textbf{(a) RoNIN, 200 Hz, 2D}\par\vspace{2pt}
\scriptsize\begin{tabular}{@{}p{1.62in}cc@{}}
\toprule
Training augmentation & Seen ATE & Unseen ATE \\
\midrule
Native RoNIN (native augmentation) & 7.08 & 10.48 \\
RoNIN-LTCA, no rotation perturbation & 8.10 & 11.33 \\
RoNIN-LTCA + independent per-step rotation perturbation & 6.14 & 9.10 \\
\bottomrule
\end{tabular}
\vspace{6pt}

\textbf{(b) TLIO, 200 Hz, 3D}\par\vspace{2pt}
\scriptsize\begin{tabular}{@{}p{1.62in}cc@{}}
\toprule
Training augmentation & GT att. & Open-loop att. \\
\midrule
Native TLIO & 2.36 & 2.42 \\
TLIO-LTCA, no rotation perturbation & 2.15 & 2.66 \\
TLIO-LTCA + independent per-step rotation perturbation & 2.08 & 2.14 \\
\bottomrule
\end{tabular}
\end{table}

Without the rotation perturbation, RoNIN-LTCA reaches a Seen ATE of 8.10 m, higher than the 7.08 m of native RoNIN; TLIO-LTCA reaches 2.15 m with ground-truth attitude, slightly better than the 2.36 m of the native model, but rises to 2.66 m under open-loop attitude, which is worse than the native 2.42 m. The long-horizon structure does not bring a gain by itself; trained entirely with ground-truth attitude, it additionally learns a coordinate consistency. When the coordinate frames of adjacent windows disagree in the real setting, the 60 s history amplifies the error of this consistency assumption along the trajectory.

With the independent per-step rotation perturbation, RoNIN-LTCA drops to 6.14 m and 9.10 m on Seen and Unseen, and the open-loop ATE-3D of TLIO-LTCA drops to 2.14 m. The attitude $\times$ perturbation 2 $\times$ 2 comparison in the TLIO panel pinpoints more directly where the effect acts: without perturbation, the ATE-3D under ground-truth and open-loop attitude differ by 0.51 m (2.15 versus 2.66 m); with perturbation, the gap shrinks to 0.06 m (2.08 versus 2.14 m). The perturbation does not level the two attitude conditions by improving single-step displacement accuracy; it breaks the reliance on coordinate consistency that the long-horizon structure develops, so that the structure can also work on datasets with sufficient excitation.

\subsection{Inference Cost}
\label{sec:4.6}

Table~\ref{tab:8} reports the inference cost of the complete architecture. Taking the TLIO backbone as an example, LTCA raises the parameter count from 5.03 M to 5.66 M (+12.5\%) and the per-step computation from 20.29 to 21.67 MMACs (million multiply-accumulate operations, +6.8\%).

\begin{table}[!t]
\centering
\caption{Per-step inference cost (TLIO backbone, batch size 1, FP32). Latency was measured on an Apple M4 CPU; the RoNIN-LTCA+Baro instance takes 3.39 / 3.67 ms. With the KV cache enabled, a step computes only the current token, and the latency is independent of the history length.}
\label{tab:8}
\scriptsize
\begin{tabular}{@{}lcc@{}}
\toprule
Metric & TLIO-LTCA & Increase over native TLIO\\
\midrule
Parameters & 5.66 M & +0.63 M (+12.5\%) \\
Per-step computation (MMACs) & 21.67 & +1.38 (+6.8\%) \\
Per-step latency (ms, mean/p95) & 2.75 / 3.12 & --- \\
\bottomrule
\end{tabular}
\end{table}

At a 1 Hz integration rate, the per-step latency of TLIO-LTCA+Baro and RoNIN-LTCA+Baro is 2.75 ms and 3.39 ms respectively, far below the 1 s period, which leaves ample margin for deployment. As an extreme reference, running two layers of causal self-attention directly over 6000 raw samples costs about 18.4 G multiply-adds for the attention matrices alone, roughly 850 times the streaming per-step cost.

\section{Discussion and Conclusion}
\label{sec:5}

\subsection{Discussion}
\label{sec:5.1}

Our experiments support one basic judgment: the difficulty of transport segments in multistory buildings is that the motion state is unobservable within a local window. The degradation of short-window models during constant-velocity cruise agrees with the observation that the specific force degenerates into constant gravity once gait impact disappears; once the onset event provides the initial state of speed and direction, the following tens of seconds only require the causal history to maintain that state. The history truncation and event masking experiments of Section~\ref{sec:4.3} provide direct evidence: when the visible history covers the onset event, the output accumulates along the true trajectory; once the event leaves the window, the state is lost and the output returns to the conditional mean of the training set; and attention concentrates on a small number of tokens near the onset event. The long-horizon structure therefore maintains the state rather than replacing the local backbone to re-observe velocity.

Compared with existing methods, LTCA leaves the observational capacity of the local backbone unchanged and only changes how state is transmitted over time. Short-window models are limited by their receptive field and cannot carry the onset event across an entire cruise; bidirectional long-window models such as IDOL enlarge the input range but require the whole window and provide no strictly causal streaming state recursion; recurrent networks maintain a continuous hidden state but lack explicit control over what history to select and forget. LTCA constrains the state to the most recent 60 s with bounded causal attention and a fixed-length cache, so that the error does not accumulate without bound along the whole trajectory and continuous 1 Hz output remains straightforward.

The design boundary also needs to be stated. A 60 s window suffices here, since elevator and escalator runs last no longer than 24 s and 45 s, but a longer transport segment would push the onset event out of the window and the state would be lost again. The value is a fixed architectural parameter rather than a per-dataset setting, since the same 60 s is used on MSIO and on both public benchmarks (Table~\ref{tab:6}); it should be chosen from the longest single transport run expected in the target scene rather than set to an arbitrary constant. The pure inertial version produces erroneous outputs during stationary segments in some unseen domains, and a few long sequences show vertical drift; both indicate that state continuity can also be triggered by spurious local features. The training-time barometric constraint alleviates this noticeably but does not remove it, and the boundary depends on the quality and availability of the barometric observation.

From an application perspective, the value of LTCA lies in upgrading an existing local backbone with a small modification: it does not change the input/output interface of the backbone, does not require changes to the attitude estimation pipeline, and its per-step cost is independent of trajectory length, which makes 1 Hz operation on a phone or an embedded device practical; the pluggable barometric module lets the system fall back to the pure inertial path when the sensor is interrupted. The difference between the two backbones also suggests that the backbone should be chosen according to the deployment domain: the TLIO backbone is more accurate in-domain, and the RoNIN backbone is more stable across domains.

\subsection{Limitations and Future Work}
\label{sec:5.2}

The conclusions of this paper are bounded by the data, the training protocol, and the scale of the evaluation. The main limitations are as follows.

\begin{enumerate}
\item \textbf{Boundary of the history window.} A fixed 60 s window requires the visible history to cover the onset of the transport segment; when a single run outlasts the window, the onset event leaves the visible range and the state is lost again. Window management for longer scenes, such as adaptive windows or event-triggered memory, remains to be studied.
\item \textbf{Cross-domain degradation and backbone differences.} LTCA instances without the barometric branch still degrade in unseen domains, mainly because of vertical drift in a few long sequences and spurious motion during gait-free stationary segments; the training-time barometric condition alleviates but does not eliminate this. The two backbones degrade by different amounts across domains (Tables~\ref{tab:2} and~\ref{tab:3}), and more unseen buildings and domain augmentation need to be studied.
\item \textbf{Sufficiency of the evaluation.} The main results come from a single training run and a single evaluation, and multi-seed means and variances are not reported; repeated training and cross-validation are needed to quantify the stability of the conclusions.
\end{enumerate}

\subsection{Conclusion}
\label{sec:5.3}

This paper extends learned pedestrian inertial odometry from the 2D plane to multistory buildings that involve vertical elevators and escalators. To address the unobservability of velocity during the cruise of passive transport, where gait excitation is absent, we propose LTCA, a long-horizon structure that can be stacked on an existing local backbone: 1 s local token pooling compresses the high-rate inertial measurements, bounded causal attention with a relative temporal bias sustains within a 60 s history the motion state established at the onset event, and 1 Hz streaming output keeps the per-step latency at the millisecond level. Experiments on MSIO, our benchmark of nine buildings, and on the public RoNIN and TLIO benchmarks show that, with the barometer online, the pluggable models on the two backbones reach 3D ATE of 2.83/4.18 m and 2.44/5.12 m on Seen and Unseen, respectively; the pure inertial version reduces the vertical error of Seen elevator segments from 2.79/3.76 m to 0.43/0.63 m and, with the barometer connected, to 0.39/0.27 m, and falls back to pure inertial inference when the sensor is interrupted, where paired conditional training still keeps the fallback inertial version above the separately trained pure inertial model. The hierarchical tokenization of LTCA allows long-range causal context to run as a low-cost stream, and its ability to establish and maintain historical state can be transferred to other environments and scenes that contain under-excited segments.

\end{document}